\documentclass{egpubl}
\usepackage{pg2023p}

\WsPaper           % uncomment for final version of EG Workshop contribution
\usepackage[T1]{fontenc}
\usepackage{dfadobe} 
\usepackage{amssymb}

\usepackage{cite}  % comment out for biblatex with backend=biber
\BibtexOrBiblatex
\electronicVersion
\PrintedOrElectronic
\ifpdf \usepackage[pdftex]{graphicx} \pdfcompresslevel=9
\else \usepackage[dvips]{graphicx} \fi

\usepackage{egweblnk} 
\title[Sketch-to-Architecture: Generative AI-aided Architectural Design Ideation]%
      {Sketch-to-Architecture: Generative AI-aided Architectural Design}
\vspace{-1em}
\author[Li et al.]{%
  Pengzhi Li$^{1}$, Baijuan Li$^{2}$, Zhiheng Li$^{1}$\thanks{Corresponding author.}\\
    $^1$ Shenzhen International Graduate School, Tsinghua University, Shenzhen, China\\
    $^2$ Shenzhen University, Shenzhen, China
}

\begin{document}

% uncomment for using teaser
% \teaser{
%  \includegraphics[width=\linewidth]{eg_new}
%  \centering
%   \caption{New EG Logo}
% \label{fig:teaser}
%}

\maketitle
%-------------------------------------------------------------------------
% \vspace{-2em}
\begin{abstract}
Recently, the development of large-scale models has paved the way for various interdisciplinary research, including architecture. By using generative AI, we present a novel workflow that utilizes AI models to generate conceptual floorplans and 3D models from simple sketches, enabling rapid ideation and controlled generation of architectural renderings based on textual descriptions. Our work demonstrates the potential of generative AI in the architectural design process, pointing towards a new direction of computer-aided architectural design. Our project website will be available at: \color{blue}{https://zrealli.github.io/sketch2arc}
\color{black}
%-------------------------------------------------------------------------

%The tool at \url{http://dl.acm.org/ccs.cfm} can be used to generate
% CCS codes.
%Example:

% \ccsdesc[300]{Computing methodologies~Collision detection}
% \ccsdesc[300]{Hardware~Sensors and actuators}
% \ccsdesc[100]{Hardware~PCB design and layout}


\end{abstract}  
%-------------------------------------------------------------------------
% \section{Introduction}

% Please follow the steps outlined in this document very carefully when
% submitting your manuscript to Eurographics.
% You may as well use the \LaTeX\ source as a template to typeset your own
% paper. In this case we encourage you to also read the \LaTeX\ comments
% embedded in the document.

\begin{figure}[t]
    \centering
    \includegraphics[width=0.48\textwidth]{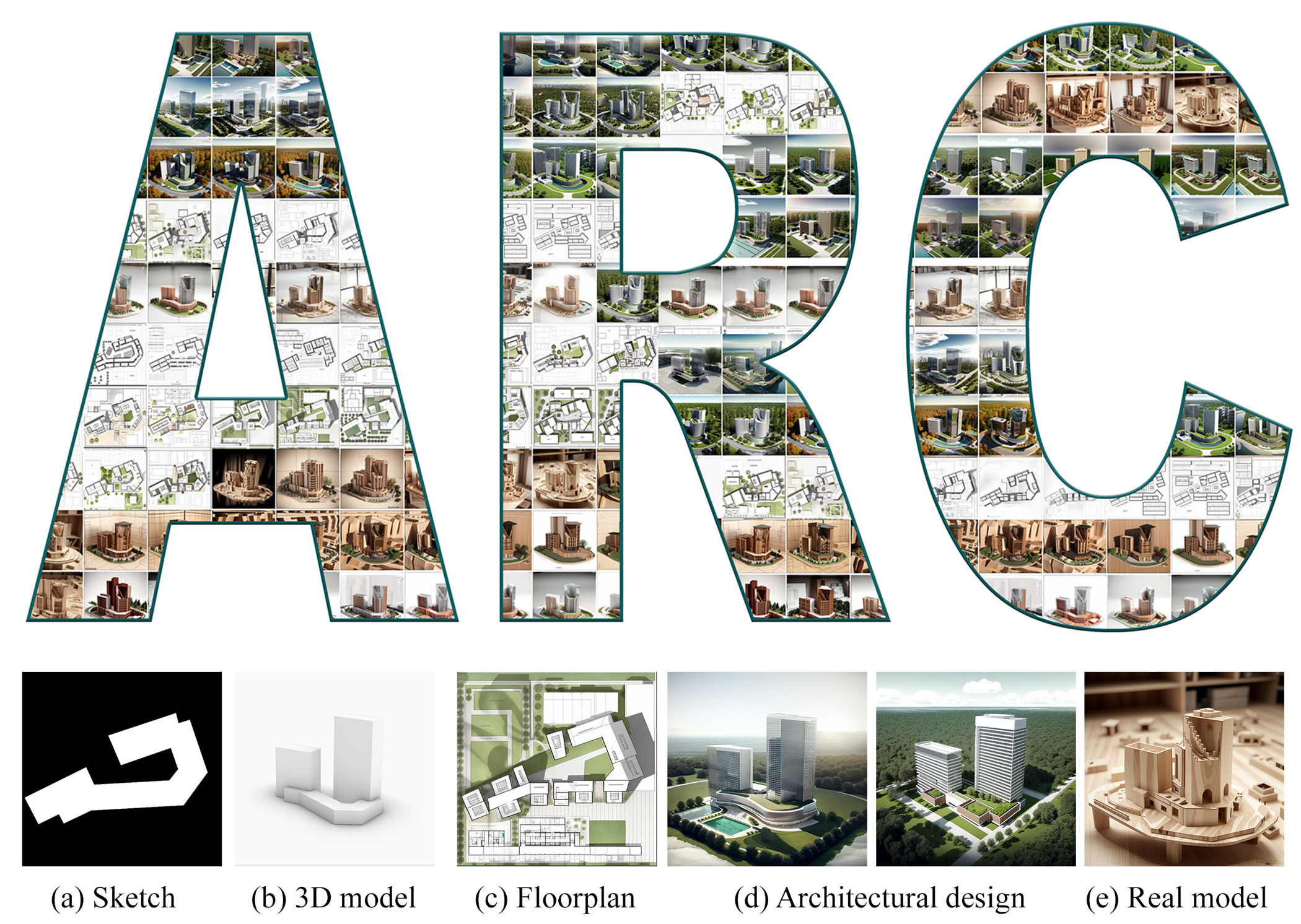}   
    \caption{We explore how generative AI technology can be effectively utilized in the early phases of architectural design. }
  \vspace{-2em}
    \label{fig:visualization1}
\end{figure}

% \vspace{-5em}
\section{Introduction}

Recently, with the rapid development of large-scale models~\cite{01ho2020denoising,02song2020denoising,03rombach2022high,04saharia2022photorealistic}, content generation methods are gradually applied across various academic disciplines. The new technologies allows us to reshape the traditional design approach of the past. Architecture, an interdisciplinary field that combines engineering and art, is characterized by its intricate and complex nature, leading to extended design cycles and uncertainties. Consequently, the preliminary architectural design phase has become complicated engineering. Traditionally, architects initiate the design process with sketches and employ visual information such as images and symbols to inspire creativity and facilitate design progress. This iterative process is time-consuming and resource-intensive. However, generative AI technology~\cite{03rombach2022high,05abdal2019image2stylegan,ruiz2023dreambooth,gal2022image,li2023layerdiffusion} has introduced new possibilities, particularly in computer-aided architectural design. This technology demonstrates significant potential in architectural design.

In this paper, as shown in Figure 1, we present a comprehensive workflow for the preliminary stages of architectural design that has not been explored previously. We explore how generative AI technology can be effectively utilized in the early phases of architectural design to generate conceptual plans and 3D models based on initial sketches. By leveraging this information, we can rapidly conceive and develop creative ideas. Additionally, we employ text-to-image generation techniques to achieve controlled generation and editing of architectural rendering images. Throughout our entire system, we can generate a range of visual cues, including architectural plans, elevations, handcrafted model, and architectural renderings, which serves as a wealth of inspiration and provide architects with many creative prompts. Our approach significantly reduces the time required for the initial stages of architectural design while offering boundless possibilities for designers' creativity. To the best of our knowledge, this is the first systematic presentation of a complete generative AI-guided workflow for the preliminary stages of architectural design. Our study reshapes the architectural design process and suggests new directions for the advancement of architectural design through state-of-the-art computer technology.

\begin{figure*}[t]
    \centering
    \includegraphics[width=0.75\textwidth]{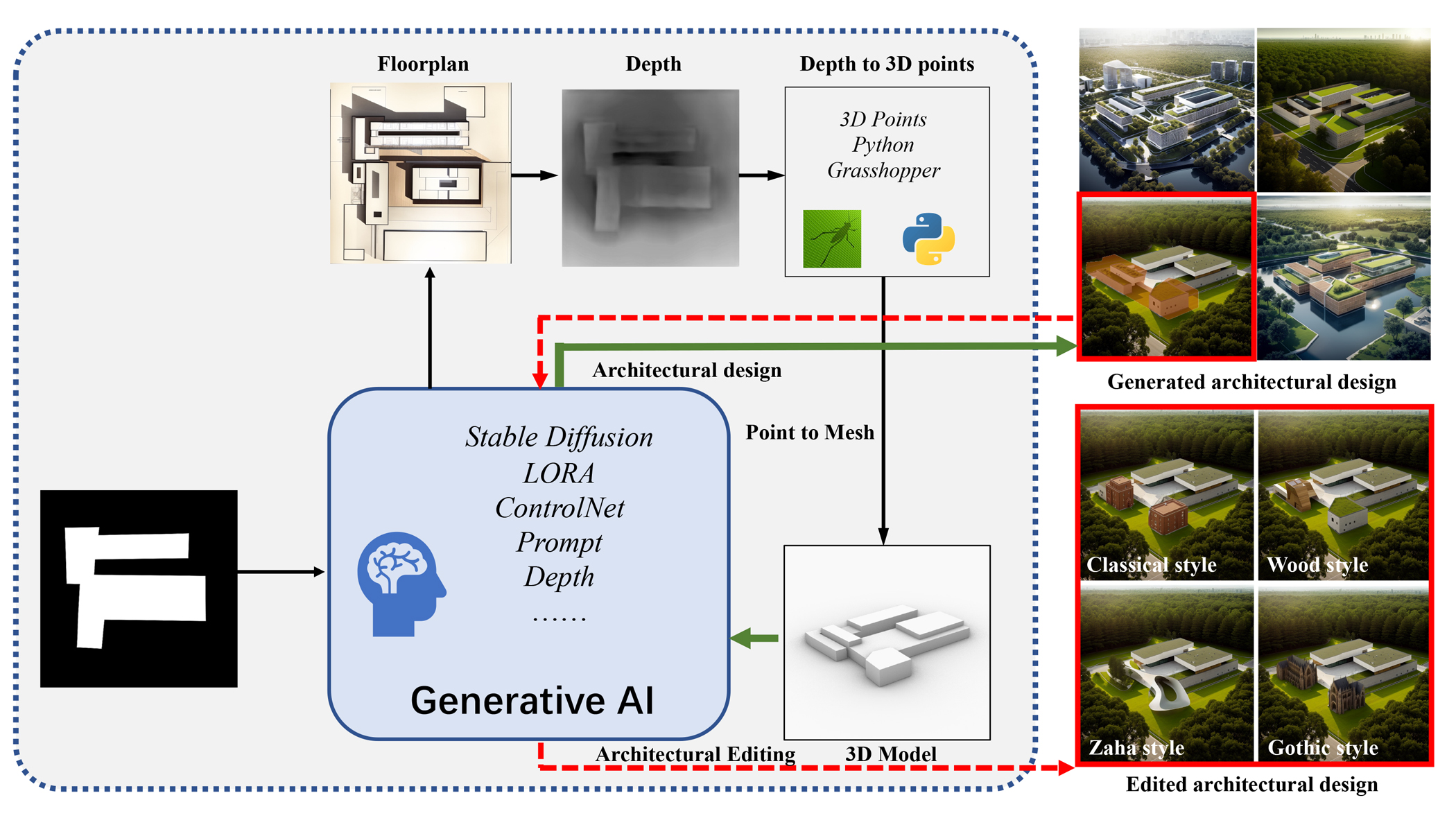}   

    \caption{We present the workflow of AI-generated architectural design. }
 \vspace{-1em}
    \label{fig:visualization1}
\end{figure*}

\section{Related works}
\subsection{Architecture generation tools}
In architectural design, Dynamo is a popular visual programming tool used for generating, analyzing, and optimizing architectural models. Developed by Autodesk, it integrates with other architectural design software, such as Revit, providing an intuitive way to control and automate the design process. Dynamo is based on graphical programming, where predefined nodes and custom scripts are connected to create complex design algorithms and workflows. These nodes can perform various functions, including creating and editing geometric shapes, manipulating and transforming data, adjusting and optimizing parameters, and more. Users can connect nodes using a simple drag-and-drop interface and adjust their behavior through parameter adjustments. By utilizing Dynamo, architects, and designers can explore the design space more flexibly, rapidly generate multiple design options, and evaluate and compare different design solutions through real-time parameter adjustments. 

Grasshopper is a widely used visual programming tool in architectural design. It allows designers to create complex architectural models and explore different design solutions by adjusting parameters in real-time through a node-based graphical programming approach. By connecting various functional nodes, designers can build custom design algorithms and workflows to accomplish tasks such as parametric design, process automation, and optimization. Grasshopper seamlessly integrates with Rhino 3D software, enabling designers to directly manipulate geometric models in Rhino and providing advanced Building Information Modeling (BIM) capabilities. Its flexibility and scalability make it an essential tool for designers to explore innovative designs, improve efficiency, and enhance design quality. Using Grasshopper, designers can expedite the design process, generate multiple design options, and optimize design solutions through parameter adjustments and optimizations, leading to more efficient, innovative, and sustainable architectural designs.
\subsection{AI in architectural design}
Recently, significant breakthroughs have been achieved in generating designs and creating incredible images. These new technologies allow architects and designers to improve existing design methodologies. Generative Adversarial Networks (GANs)~\cite{07karras2019style,08karras2020analyzing} have been proven successful in the past few years, allowing designers to generate various architectural floorplans based on user input. However, their focus is primarily on floorplans~\cite{09nauata2020house} rather than comprehensive architectural designs. Furthermore, due to the required machine learning knowledge, only some designers possess sufficient expertise to study and utilize them effectively.
The application of AI in architectural generations~\cite{10galanos2023architext,11seneviratne2022dalle,12as2018artificial} is growing, with the most common being direct image generation from textual descriptions. Text-to-image generation methods offer an accessible approach that generates architectural design images based on text prompts. However, the current process of text-to-image generation needs more control. To control the output, users not only need to employ specific keywords to guide the generation of images with desired styles but also face challenges in controlling the geometric details of the architectural generation. Further exploration is still needed to refine its application in architecture.

\begin{figure}[t]
    \centering
    \includegraphics[width=0.49\textwidth]{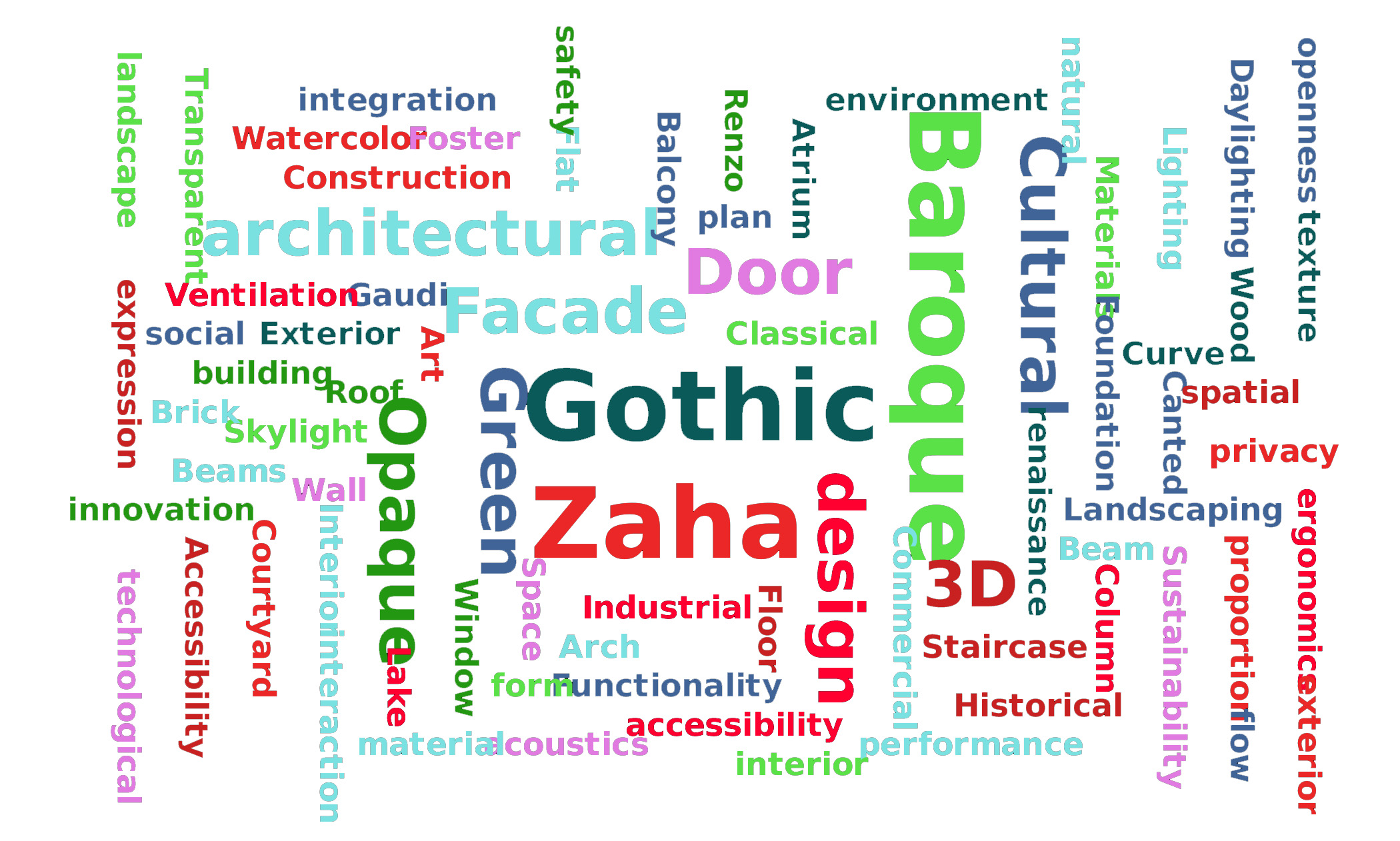}   

    \caption{We show some of the most critical terms in architectural design, covering various architectural styles, architectural types, architectural materials, etc.}
 
    \label{fig:visualization1}
\end{figure}

\begin{figure}[t]
    \centering
    \includegraphics[width=0.49\textwidth]{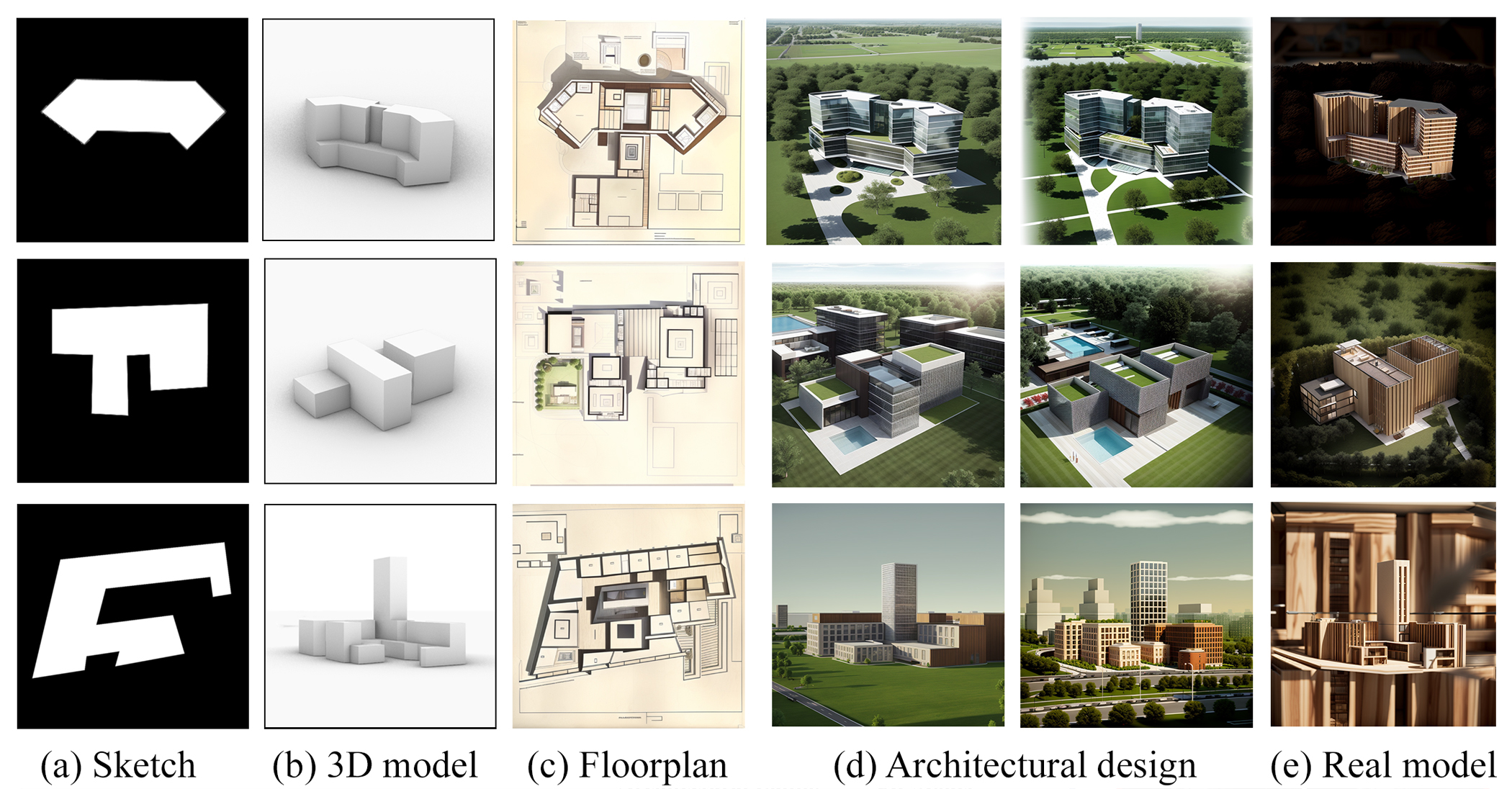}   

    \caption{We show more architectural designs generated from simple sketches by our method.}
 \vspace{-2em}
    \label{fig:visualization1}
\end{figure}

\begin{figure*}[t]
    \centering
    \includegraphics[width=0.8\textwidth]{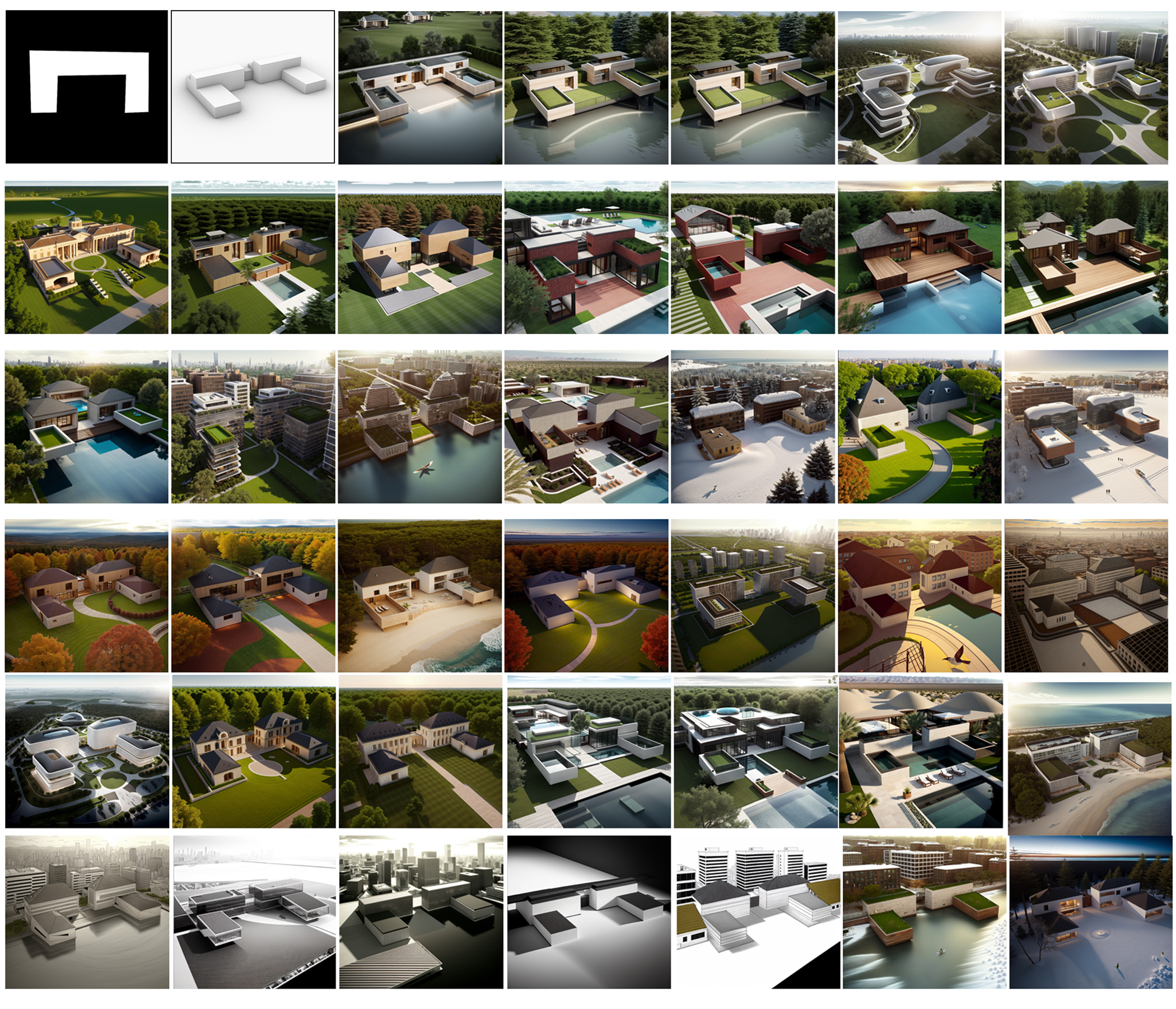}   

    \caption{We show many architectural design results. The results are generated based on different architectural design terms, including architectural style and types. More textual details can be found in Figure 3.}
 
    \label{fig:visualization1}
\end{figure*}

\section{Method}
\subsection{Preliminaries}
The Stable Diffusion model~\cite{03rombach2022high} represents a significant milestone in the advancement of generative AI due to its high-performance architecture, which produces high-quality images. This model primarily focuses on generating images based on text prompts (text-to-image) and facilitates image editing using both text and image inputs.
Low-Rank Adaption of large language model (LoRA)~\cite{13hu2021lora} is a fine-tuning method specifically tailored for large-scale language models. It involves keeping the parameters of the large network fixed while training only specific layers, such as specific linear layers (e.g., the linear projection of QKV in transformers~\cite{14vaswani2017attention} and the linear layers of FFN). This approach significantly reduces the number of training parameters during the fine-tuning process.
ControlNet~\cite{15zhang2023adding} plays a crucial role in fine-tuning with a small dataset. Its objective is to ensure that the trained model effectively adapts to the specific dataset while minimizing any impact on the original model's capabilities.

\subsection{Floorplan and 3D massing model generation}

The overall workflow is described in Figure 2, our approach is based on the state-of-the-art  Stable Diffusion model~\cite{rombach2022stable} and involves fine-tuning it for different design tasks. To achieve this, we collect numerous images of architectural floorplans encompassing various categories, such as residential, commercial, and museum designs. We utilize different kinds of floorplan data for fine-tuning the LoRA~\cite{hu2021lora} model, specifically adapted to different types of architectural design tasks. LoRA~\cite{hu2021lora} injects trainable layers into each transformer block instead of using pre-trained model weights, significantly reducing the number of trainable parameters and resulting in faster computation and lower computational requirements while achieving comparable results to those obtained by fine-tuning the entire large model. Additionally, we incorporate the recently released ControlNet~\cite{15zhang2023adding} model to achieve more precise conditional control in generating floorplans with the diffusion model. 

After generating the floorplans, we utilize the depth estimation model~\cite{16ranftl2020towards} to predict the depth maps of the floorplans, which are then converted into grayscale images. Subsequently, the grayscale images are fed into Rhino to construct 3D point cloud models. We transform the point cloud models into 3D meshes using mesh scripts developed in Grasshopper, resulting in the final architectural massing models. As a result, we can generate architectural elevations from multiple perspectives. 
% More architectural design results are shown in Figure 3.

\vspace{-1em}
\subsection{Architectural design}

 In the past, parametric designers had to develop multiple architectural algorithms based on building rules and combine them to obtain desired results~\cite{17buhamdan2021generative,18baduge2022artificial}. This approach made the entire early architectural design process complicated. However, our proposed method enables controlled architectural generation and editing through text prompts. We first extract core terms from conceptual architectural design, which are crucial in determining the final architectural design.
 The core elements are in Figure 2 of the supplementary materials, including building types, architectural styles, architectural features, building materials, architectural perspectives, and more. A comprehensive early architectural design must encompass these visual information elements, and our method revolves around this series of design elements.
 From a technical perspective, our method still employs the fine-tuning approach of the diffusion model, utilizing the simultaneous input of text and images to generate entirely new results that align with specific core elements.

However, the design will undergo iterations and revisions in practical architectural design. To address this, we utilize masks to modify specific regions of the generated building, such as material editing, element modifications, and structural changes. The details of local editing can be found in the supplementary materials. More architectural design results are shown in Figure 5.

\section{Results}
Firstly, we generate floorplans and architectural 3D massing models based on the sketches. Then, we employ the generated massing models to achieve the end-to-end generation of architectural renderings. This process is controlled by fine-tuned models and textual descriptions, allowing us to integrate various design requirements and obtain the desired preliminary architectural design results. Figure 4 illustrates the overall generated results. The textual descriptions are composed using the architectural design elements in the supplementary material.

\begin{figure*}[t]
    \centering
    \includegraphics[width=0.8\textwidth]{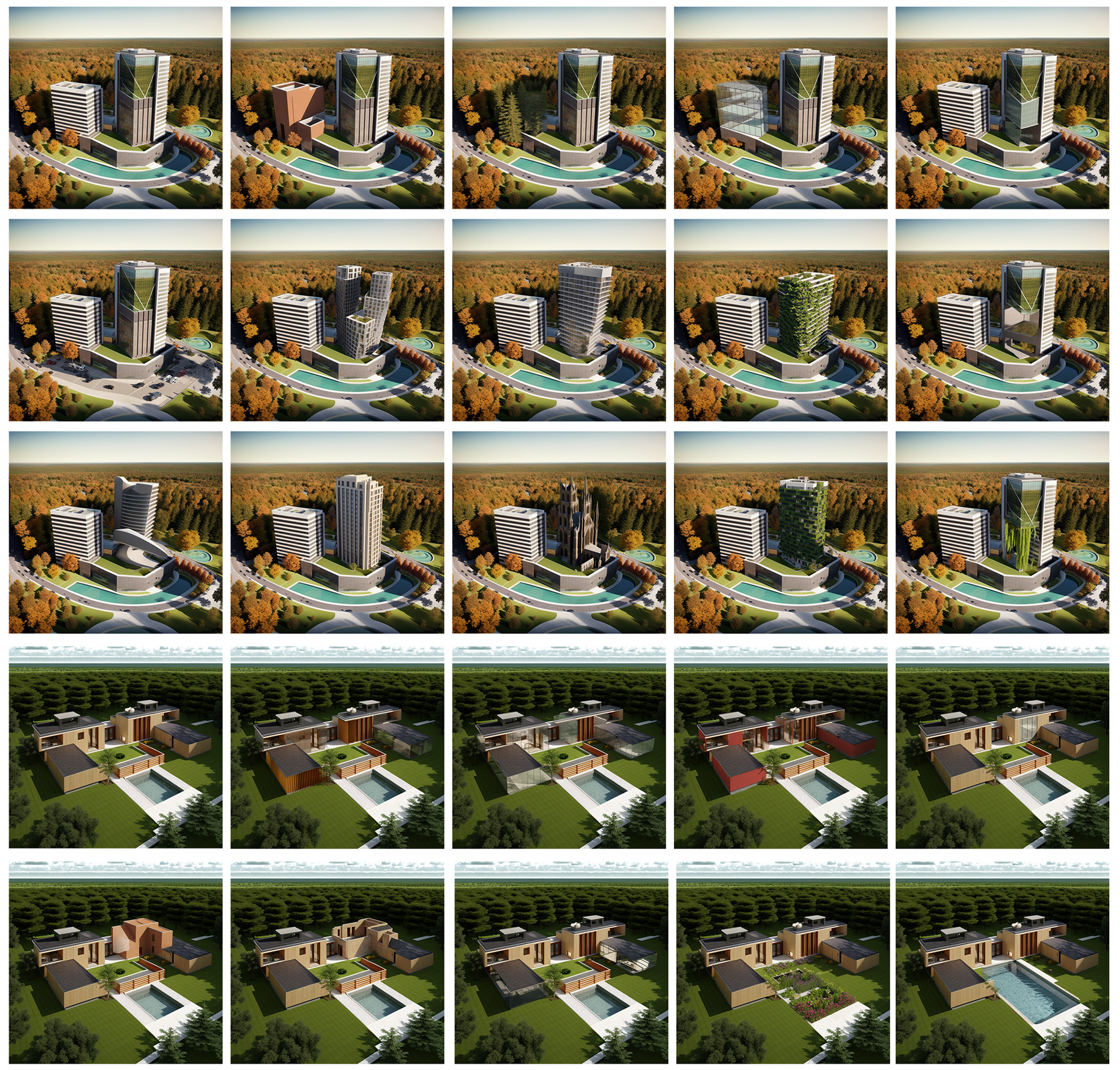}   

    \caption{The visualization results of architectural editing. By utilizing masks, specific regions of the generated building are modified, including material editing, element modifications, enabling real-time targeted edits while preserving the overall architectural design.}
 
    \label{fig:visualization2}
    \vspace{-1em}
\end{figure*}

\section{Visualization}
\subsection{Global generation}
Firstly, we generate floorplans and architectural 3D models based on the sketches. Then, we employ the generated architectural models to achieve the end-to-end generation of architectural renderings. This process is controlled by fine-tuned models and textual descriptions, allowing us to integrate various design requirements and obtain the desired preliminary architectural design results. Figure 5 illustrates the overall generated results. The textual descriptions are composed using the architectural design elements in Figure 3.

\textit{Architectural styles:} Architects learn about various design styles and research the architectural styles of their preferred designers. With the existing construction techniques, most design concepts can be adequately realized. In order to enable AI to adapt to this task, we collect kinds of architectural style images. By fine-tuning the diffusion model, we train the model to learn diverse architectural styles and the individual styles of famous architects, including Baroque, Postmodern, and the works of renowned architects like Zaha Hadid and Frank Gehry. Our method enables the generation of architectural designs in specific styles based on sketches.

\textit{Architectural types:}The appearance of buildings often varies according to their different functions. In the field of architecture, the design of a building shows significant differences based on its intended purpose and type. We conduct experiments for generating facades and renderings of different types of buildings by utilizing various textual cues such as commercial, residential, and other relevant keywords. 
Architectural constructions: We can explore and construct more intricate architectural forms with significant advancements in design and manufacturing technologies. It enables architects to challenge conventional design constraints and create unique buildings with striking visual impact. We research and experiment with various architectural forms, such as circular and sloping structures, to showcase their potential and application in architectural design.

\textit{Architectural materials:} Architectural materials also play a crucial role in design. They determine the appearance, texture, and performance of a structure. Appropriate material selection can create a unique style and provide necessary functional capabilities while considering sustainability and environmental conservation. 

\textit{Architectural landscape: }Architectural landscaping plays a significant role in architectural design. It encompasses various elements of the surrounding environment, such as land, vegetation, water bodies, artificial structures, and their relationship with the buildings. Architectural landscaping design aims to create an environment that complements and integrates with the architecture. 

\textit{Architectural renderings:} Architectural rendering is a process that transforms design concepts into realistic images and animations, serving as a powerful tool for effectively communicating the creative ideas of designers to clients, stakeholders, and decision-makers. By showcasing various aspects, such as the building's appearance, proportions, materials, and integration with the environment, architectural rendering helps people better understand and experience architectural designs. It enables designers to accurately convey their design intentions.

\subsection{Local editing}
Based on those above key architectural terms in Figure 3, we can obtain initial visualization results for an architectural generation. However, the design will undergo multiple iterations and revisions in practical architectural concept design. To address this, we utilize masks~\cite{avrahami2023blended} to modify specific regions of the generated building, such as material editing, element modifications, and structural changes. Our approach is close to the parametric design~\cite{19monedero2000parametric,20schnabel2007parametric}. However, instead of using numerical parameters, we employ text prompts as variables and leverage the powerful generative capabilities of large-scale models to perform specific editing operations~\cite{avrahami2023blended}, providing real-time feedback to architects. As shown in Figure 6, we extensively explore local architectural design edits based on the architectural terms discussed earlier. Our approach allows desired modifications to specific elements while preserving the rest of the architectural rendering unchanged.

\vspace{-1em}

\section{Future work}
The emergence of AI design tools is similar to the digital modeling tools(e.g., AutoCAD and Rhino) and currently only serves as aids to traditional architectural design workflows. In this context, it is essential to recognize the changes AI technology brings to the design process. Our work shows a potential new design workflow primarily focused on early-stage architectural conceptual design. However, this approach brings several issues, including the requirement for enhanced functionality in floorplans, addressing imbalanced proportions, and resolving the inconsistency in multiple perspectives within renderings. In fact, due to the unique development of interdisciplinary approaches, these issues will inevitably be addressed by optimized models tailored to architectural design. Our future task is to incorporate more architectural design standards into model training, thus generating more reliable architectural design solutions that adhere to building regulations.

\vspace{-1em}
\section{Conclusion}
This paper presents a method for achieving a rapid and creative workflow in the early stages of architectural design using generative AI technology. We rapidly generate architectural solutions from simple sketches by combining diffusion models with deep estimation models and scripts written in Rhino software. Additionally, we propose a controllable architectural generation method, allowing designers to control the architectural design by inputting specific text prompts. Our novel approach significantly improves design efficiency and enhances design quality. This is the first systematic presentation of a complete architectural early-stage design guided by generative AI. Our work reshapes the architectural design process, pointing toward new directions in architectural design development on the threshold of new and emerging technologies.

\bibliographystyle{eg-alpha-doi}

\bibliography{egbibsample}

%-------------------------------------------------------------------------
% \newpage

\end{document}